# Language, logic and ontology: uncovering the structure of commonsense knowledge[*]

**Walid S. Saba**

*Computational and Statistical Sciences Center*
*American Institutes for Research*
*1000 Thomas Jefferson Street, NW,*
*Washington, DC 20007*
*WSaba@air.org*

**Abstract**

The purpose of this paper is twofold: (*i*) we argue that the structure of commonsense knowledge must be discovered, rather than invented; and (*ii*) we argue that natural language, which is the best known theory of our (shared) commonsense knowledge, should itself be used as a guide to discovering the structure of commonsense knowledge. In addition to suggesting a systematic method to the discovery of the structure of commonsense knowledge, the method we propose seems to also provide an explanation for a number of phenomena in natural language, such as metaphor, intensionality, and the semantics of nominal compounds. Admittedly, our ultimate goal is quite ambitious, and it is no less than the systematic 'discovery' of a well-typed ontology of commonsense knowledge, and the subsequent formulation of the long-awaited goal of a meaning algebra.

***Keywords***: *Ontology, semantics, commonsense knowledge, reasoning*

## 1. Introduction

In *Logic and Ontology* Cocchiarella (2001) convincingly argues for a view of "logic as a language" in contrast with the (now dominant) view of "logic as a calculus". In the latter, logic is viewed as an "abstract calculus that has no content of its own, and which depends on set theory as a background framework by which such a calculus might be syntactically described and semantically interpreted." In the view of "logic as a language", however, logic has content, and "ontological content in particular." Moreover, and according to Cocchiarella, a logic with ontological content necessitates the use of type theory (and predication), as opposed to set theory (and set membership), as the background framework. An obvious question that immediately comes to mind here is the following: what exactly is the nature of this strongly-typed ontological structure that will form the background framework for a new logic that has content?

In our opinion, part of the answer lies in an insightful observation that Hobbs (1985) made some time ago, namely that difficulties encountered in the semantics of natural

---

language are due, in part, to difficulties encountered when one attempts to specify the exact nature of the relationship between language and the world. While it has not received much attention, the crucial point that Hobbs makes is the observation that if one "assumes a theory of the world that is isomorphic to the way we *talk* about it" (emphasis added), then "semantics becomes very nearly trivial". The picture we have in mind, depicted graphically in figure 1, is a logic and a semantics that is grounded in a strongly-typed ontology, an ontology that in turn reflects our commonsense view of the world and the way we talk about it.

Assuming the existence of such an ontological structure, semantics might indeed become 'nearly' trivial, and this is demonstrated in this paper by investigating some challenging problems in the semantics of natural language, namely lexical ambiguity, the semantics of intensional verbs and the semantics of nominal compounds.

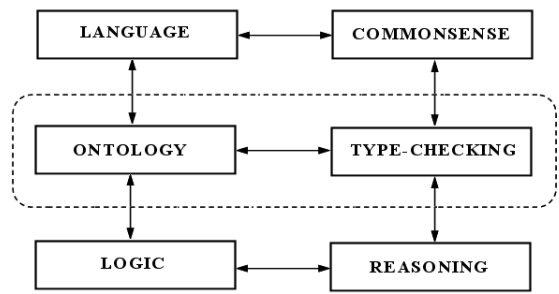

**Figure 1.** Language, logic and ontology

In the remainder of this paper we (*i*) discuss the intimate relationship between language and knowledge and argue that language 'understanding' necessitates the use of a strongly-typed ontological structure that reflects our commonsense view of the world; (*ii*) we briefly outline a process that uses language itself as a guide to discovering the nature of this ontological structure; (*iii*) we show how the semantics of several natural language phenomena becomes nearly trivial in a logic that is grounded in an ontological structure that is isomorphic to our commonsense view of the world; and (*iv*) we finally conclude by discussing some steps towards achieving the long-awaited dream of a meaning algebra.

## 2. Language and Knowledge

Cognitive scientists have long recognized the intimate relationship between natural language understanding (NLU) and knowledge representation and reasoning (KR&R) - or, in short, the intimate relationship between language and knowledge. In fact, research in NLU seems to have been slowly embracing what we like to call the 'understanding as reasoning' paradigm, as it has become quite clear by now that understanding natural language is, for the most part, a commonsense reasoning process at the pragmatic level. As an example illustrating this strong interplay between language understanding and commonsense reasoning, consider the following:

(1) a) *john defended a jailed activist in every country*
    b) *john knows a jailed activist in every country*

From the standpoint of commonsense, most readers would find no difficulty in a reading for (1*a*) that implies John's support for the 'same' activist in every country. However, the same is not true in (1*b*), as one can hardly conceive of a single activist being jailed in every country. Thus, while a wide scope *a*, implying a single activist is quite plausible in (1*a*), the more plausible reading in (1*b*) is the one implying several activists, making (1*b*) read something like 'in *every country, John knows some jailed activist*'. What we suggest here is that such inferences lie beyond syntactic and semantic explanations, and are in fact a function of our commonsense knowledge of how the world (the *possible world* we actually live in!) really is. This process is even more complex due to the fact that different individuals may have different scope preferences in the same linguistic context, as the experiments of Kurtzman & MacDonald (1993) have suggested. Consistent with this 'understanding as reasoning' paradigm, an inferencing strategy that models individual preferences in the resolution of scope ambiguities at the pragmatic level has been suggested in (Saba & Corriveau, 1997). While it has been argued that such problems do not always require the storage of and reasoning with vast amounts of background knowledge (see Saba & Corriveau, 2001), other linguistic comprehension tasks clearly do. For instance, consider the resolution of 'He' in the following:

(2) *John shot a policeman. He immediately*
    a) *fled away.*
    b) *fell down.*

Clearly, such references must be resolved by recourse to commonsense knowledge – for example, that, typically, when *Shot*($x$,$y$) holds between some $x$ and some $y$, $x$ is the more likely subject to flee (2*a*), and $y$ is the more likely subject to fall down (2*b*). Note, however, that such inferences must always be considered defeasible, since quite often additional information might result in the retraction of previously made inferences. For example, (2*b*) might, after all, be describing a situation in which John, a 7-year old who was shooting a bazooka, fell down. Similarly, (2*a*) might actually be describing a situation in which the policeman, upon being slightly injured, tried to flee away, perhaps to escape further injuries! Computationally, there are clearly a number of challenges in reasoning with un-committed (or 'underspecified') logical forms, and this has indeed received considerable attention by a number of authors (e.g., see Kameyama, 1996, and the excellent collection of papers in van Deemter & Peters, 1996). However, the main challenge that such processes still face is the availability of this large body of commonsense knowledge along with a computationally effective reasoning engine.

While the monumental challenge of building such large commonsense knowledge bases was indeed faced head-on by a few authors (e.g., Lenat & Ghua, 1990), a number of other authors have since abandoned (and argued against) the 'knowledge intensive' paradigm in favor of more quantitative methods (e.g., Charniak, 1993). Within linguistics and formal semantics, little or no attention was paid to the issue of commonsense reasoning at the pragmatic level. Indeed, the prevailing wisdom was that NLU tasks that require the storage of and reasoning with a vast amount of background knowledge were 'highly undecidable' (e.g., van Deemter, 1996; Reinhart, 1997).

In our view, both trends were partly misguided. In particular, we hold the view that (*i*) language 'understanding' is for the most part a commonsense 'reasoning' process at the pragmatic level, and, consequently, the knowledge bottleneck problem cannot be solved by being ignored, but must be faced head-on; and (*ii*) the 'understanding as reasoning' paradigm, and the underlying knowledge structures that it utilizes, must be *formalized* if we ever hope to build scalable systems (or, as John McCarthy once said, if we ever hope to build systems that we can actually understand!). In this light we believe the work on integrating logical and commonsense reasoning in language understanding (Allen, 1987; Pereira & Pollack, 1991; Zadrozny & Jensen, 1991; Hobbs, 1985; Hobbs et al., 1993; Asher & Lascarides, 1998; and Saba & Corriveua, 2001) is of paramount importance[1].

The point we wish to make here is that successful NLU programs necessitate the design of appropriate knowledge structures that reflect our commonsense view of the world and the way we talk about it. That, on its own, is not nearly novel. Indeed, investigating the formal properties of the commonsense world have long been investigated in the pioneering work of Hayes and Hobbs (1985). Moreover, a number of other substantial efforts towards building ontologies of commonsense knowledge have also been made since then (e.g., Lenat & Ghua, 1990; Mahesh & Nirenburg, 1995; Sowa, 1995), and a number of promising trends that advocate ontological design based on sound linguistic and logical foundations have indeed started to emerge in recent years (e.g., Guarino & Welty, 2000; Pustejovsky, 2001). However, a *systematic* and *objective* approach to ontological design is still lacking. In particular, we believe that an ontology for commonsense knowledge must be *discovered* rather than *invented*, and thus it is not sufficient to establish some principles for ontological design, but that a strategy by which a commonsense ontology might be *systematically* and *objectively* designed must be developed. In this paper we propose such a strategy.

## 3. Language and Ontology

Our basic strategy for discovering the structure of commonsense knowledge is rooted in Frege's conception of Compositionality. According to Frege (see Dummett, 1981, pp. 4-7), the sense of any given sentence is derived from our previous knowledge of the senses of the words that compose it, together with our observation of the way in which they are combined in that sentence. The cornerstone of this paradigm, however, is an observation regarding the manner in which words are supposed to acquire a sense that, in our opinion, has not been fully appreciated. In particular, the principle of Compositionality is rooted in the thesis that "our understanding of [those] words consists in our grasp of the way in which they may figure in sentences in general, and how, in general, they combine to determine the truth-conditions of those sentences." (Dummett, 1981, pp. 5). Thus, the meanings of words (i.e., the concepts and the corresponding ontological structure), and the relationships between them, can be reverse-engineered, so to speak, by analyzing how these words are *used* in everyday language. As will be argued below, reservations (that abound!) regarding Compositionality can be alleviated once the roots of Frege's Compositionality is understood in this light. To see this let us first begin by introducing a predicate *App(p,c)* which is taken to be true of a property *p* and a concept *c iff* "it makes sense to speak of the property *p* of *c*". Consider now the following two sets of adjectives and nouns:

(3)  *P* = {*Strong, Smart, Imminent*}
(4)  *C* = {Table, Elephant, Event}

---

[1] Outside the domain of NLU, other pioneering work such as that of (McCarthy 1980), was also done in the same spirit, namely to integrate logical and commonsense reasoning.

A quick analysis of *App(p,c)* on the nine adjective-noun combinations yields the structure shown in figure 2. That is, while it makes sense to say 'strong table', 'strong elephant', and 'smart elephant', it does not make sense to say 'smart table', 'smart event', 'imminent elephant', etc. First it must be pointed out that the structure shown in figure 2 was *discovered* and not *invented*! Moreover, in applying the predicate *App(p,c)* on a specific property/relation *p* and a specific concept *c*, we must be wary of metaphor and polysemy.

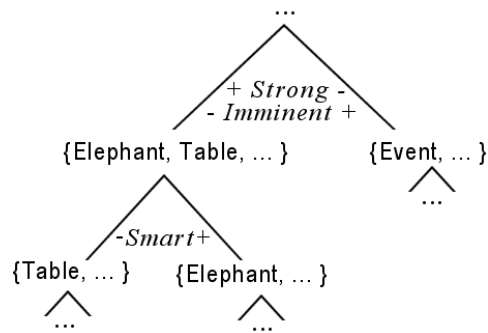

**Figure 2.** Structure resulting from the analysis of *App(p,c)* on nine adjective-noun combinations

For example, while it makes sense to say 'big table' and 'big discovery', it is clear that the sense of 'big' is different in the two instances. We will have more to say about metaphor and polysemy later in the paper. For now, however, it should be pointed out that this kind of analysis is not much different from the type inferencing process that occurs in modern, strongly-typed, polymorphic programming languages. As an example, consider the type inferences corresponding to the linguistic patterns shown in table 1.

| Pattern | Type Inference |
|---|---|
| $x + 3$ | $x$ is a Number |
| *Reverse*($x$) | $x$ is a Sequence |
| *Insert*($x,y$) | $x$ is an Thing; $y$ is Sequence of $x$ Things |
| *Head*($x$) | $x$ is a Sequence |
| *Even*($x$) | $x$ is a Number |

**Table 1.** Some linguistic patterns and their corresponding type inferences

From $x + 3$, for example, one can infer that $x$ is a number since numbers are the "kinds of things" that can be added to 3 (or, for the expression '$x + 3$' to make sense, $x$ must be a Number!) In general, the most generic type possible is inferred (i.e., these operations are assumed to be polymorphic). For example, all that can be inferred from *Reverse*($x$) is that $x$ is the generic type Sequence, which could be a List, a String (a sequence of Characters), a Vector, etc. Note also that in addition to actions (called functions or methods in programming lingo), properties (truth-valued functions) can also be used to infer the type of an object. For example, from *Even*($x$) one can infer that $x$ is a Number, since lists, sequences, etc. are not the kinds of objects which can be described by the predicate *Even*. This process can be more formally described as follows:

1. we are given a set of concepts $C = \{c_1,..., c_m\}$ and a set of actions (and properties) $P = \{p_1,..., p_m\}$
2. a predicate *App(p,c)*, where $c \in C$ and $p \in P$ is said to be true *iff* the action (or property) *p* applies to (makes sense of) objects of type *c*.
3. a set $C_p = \{c \mid App(p,c)\}$, denoting all concepts *c* for which the property *p* is applicable is generated, for each property $p \in P$.
4. a concept hierarchy is then systematically discovered by analyzing the subset relationship between the various sets generated.

To illustrate how this process (systematically) yields a type hierarchy, we consider applying the predicate *App(p,c)*, where $p \in P$ and $c \in C$ and where the sets *C* and *P* are the following:

*C* = {Set, List, Bag, Map, Tree, Relation}
*P* = {*Reverse, Size, MemberOf, Head, Tail, ElementAt, NumOfOccur, RemoveDups, Root, Leaves, etc.*}

The result of repeated analysis of the predicate *App(p,c)* on these sets results in the following:

$C_{Size} = \{Set, Bag, Relation, Map, Tree, List\}$

$C_{NumOfOccur} = \{Bag, List\}$

$C_{MemberOf} = \{Set, Bag, List, Map, Tree\}$

$C_{RemoveDups} = \{Bag, List\}$

$C_{Reverse} = \{List\}$

*etc.*

Thus, whatever the type of the container, we can always ask whether a certain object is a *MemberOf* the container; whether or not the container is *Empty*; and it always makes sense to ask for the *Size* of a container. Moreover, and while it makes sense to ask for the number of occurrences (*numOfOccur*) of a certain object in both a Bag and a List (as both can contain duplicates), it only makes sense to speak of the *Head* and *Tail* of a List, as the order of the objects in a Bag is not important. Thus, a Bag is a Set with duplicates, and a List is a Bag (and thus a Set) with duplicates and where the order is important. The result of the above analysis is the structure shown in figure 3. What is important to note here is that each (unique) set corresponds to a concept in the hierarchy.

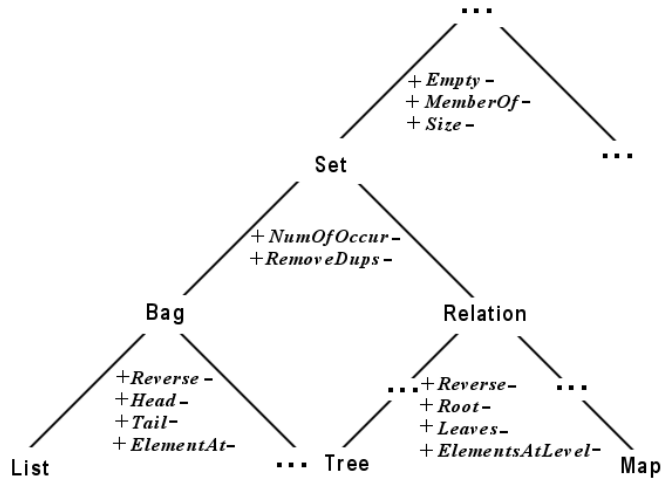

**Figure 3.** Structure implied by repeated applications of the *App(p,c)* predicate on several objects and properties

Equal sets (e.g. $C_{Tail}$ and $C_{Head}$) correspond to the same concepts. The label of a given concept could be any meaningful label that intuitively represents all the sub-concepts in this class. For example, in figure 3 Set was used to collectively refer to any collection (ordered, unordered, with or without duplicates, etc). It is also interesting to note the similarity at structurally isomorphic places in the hierarchy. For example, while we may ask for the *Head* and the *Last* of a List, we usually speak of the *Root* and the *Leaves* of a Tree. As will be discussed below, in the context of natural language, properties at structurally isomorphic locations represent metaphorical derivations, while variations in the interpretation of a property at lower levels (e.g., *Empty* in figure 3) represents polysemy. Finally, it should be noted here that there are a number of rules that can be established from the concept hierarchy shown in figure 3. For example, one can state the following:

(5) $(\forall c)(App(Reverse,c) \supset App(Size,c))$
(6) $(\exists c)(App(Size, c) \land \neg App(Reverse,c))$
(7) $(\forall c)(App(Tail,c) \equiv App(Head,c))$

Here (5) states that whenever it makes sense to reverse an object *c*, then it also makes sense to ask for the size of *c*. This essentially means that an object to which the *Size* operation can be applied is a parent of an object to which the *Reverse* operation can be applied. (6), on the other hand, states that there are objects for which the *Size* operation applies, but for which the *Reverse* operation does not apply. Finally, (7) states that whenever it makes sense to ask for the *Head* of an object then it also makes sense to ask for its *Tail*, and vice versa. It is important to note here that in performing this analysis we have assumed that *App(p,c)* is a Boolean-valued function, which has the consequence that the resulting type hierarchy is a strict binary tree. In fact, this is one of the main characteristics of our method, and has led to two important results: (*i*) multiple inheritance is completely avoided; and (*ii*) by not allowing any ambiguity in the interpretation of *App(p,c)*, lexical ambiguity, polysemy and metaphor are explicitly represented in the hierarchy. This issue will be discussed in more detail below.

## 4. Language and Commonsense Knowledge

The work described here was motivated by the following two assumptions: (*i*) the process of language understanding is, for the most part, a commonsense reasoning process at the pragmatic level; and (*ii*) since children master spoken language at a very young age, children must be performing commonsense reasoning at the pragmatic level, and consequently, they must posses all the commonsense knowledge required to understand spoken language[2]. In other words, we are assuming that deciding on a particular *App(p,c)* is not controversial, and that children can easily answer questions such as those shown in table 2 below.

    Note that in answering these questions it is clear that one has to be coconscious of metaphor. For example, while it is quite meaningful to say strong table, strong man, and strong feeling, it is clear that the senses of strong in these three cases are quite distinct. The issue of metaphors will be dealt with below. For now, all that matters, initially, is to

---

[2] Thus it may very well be the case that *everything we need to know we learned in kindergarten*!

consider posing queries such as *App*(*Smart*, Elephant) – or equivalently, questions such as 'does it make sense to say smart elephant?', to a five-year old. Furthermore we claim that *App*(*p,c*) is binary-valued; that is, while it could be a matter of degree as to how smart a certain elephant might be (which is a quantitative question), the qualitative question of whether or not it is meaningful to say 'smart elephant' is not a matter of degree[3].

| Query | Does it make sense to say… |
|---|---|
| *App*(*Walk*, Elephant) | elephants walk? |
| *App*(*Talk*, Elephant) | elephants talk? |
| *App*(*Smart*, Elephant) | elephants are smart? |
| *App*(*Scream*, Book) | books scream? |
| *App*(*Happy*, Sugar) | happy sugar? |

**Table 2.** Deciding on a particular *App*(*v,c*) from the standpoint of commonsense.

With this background we now show that an analysis of how verbs and adjectives are used with nouns in everyday language can be used to *discover* a fragment of the structure of commonsense knowledge:

1. We are given $P = \{p_1, ..., p_m\}$, a set of (distinct senses of) adjectives and verbs,
2. We are given $C = \{c_1, ..., c_n\}$, a set of (distinct senses of) nouns
3. Generate $C_i = \{c \mid App(p_i, c_j)\}$ for every pair $(p_i, c_j)$, $p_i \in P$ and $c_i \in C$
4. Generate the structure implied by all sets $C_i \in \{C_1, ..., C_m\}$

We are currently in the process of automating this process, and in particular we are planning on generating some of the sets in step 3 above by analyzing a large corpus. The fragment of the structure shown in figure 4 was however generated manually by analyzing about 1000 adjectives and verbs as to how they may or may not apply to (or make sense of) about 1000 nouns.

---

[3] As Elkan (1993) has convincingly argued, to avoid certain contradictions logical reasoning must at some level collapse to a binary logic. While Elkan's argument seemed to be susceptible to some criticism (e.g., Dubois et al (1994)), there are more convincing arguments supporting the same result. Consider the following:

(1) *John likes every famous actress*

(2) *Liz is a famous actress*

(3) *John likes Liz*

Clearly, (1) and (2) should entail (3), regardless of how famous Liz actually is. Using any quantitative model (such as fuzzy logic), this intuitive entailment cannot be produced (we leave to the reader the details of formulating this in fuzzy logic) The problem here is that at the qualitative level the truth-value of *Famous*(*x*) must collapse to either true or false, since at that level all that matters is whether or not Liz is famous, not how certain we are about her being famous.

Note that according to our strategy every concept at the knowledge- (or commonsense) level must 'own' some unique property, and this must also be linguistically reflected by some verb or adjective. This might be similar to what Fodor (1998, p. 126) meant by "having a concept is being locked to a property". In fact, it seems that this is one way to test the demarcation line between commonsense and domain-specific knowledge, as domain-specific concepts do not seem to be uniquely locked to any word in the language. Furthermore, it would seem that the property a concept is locked to (e.g., the properties *Reason* and *Talk* of a RationalLivingThing or a Human) is closely related to the notion of immutability of a feature discussed in (Sloman et al, 1998), where the immutable features of a concept are those features that collectively define the essential characteristics of a concept.

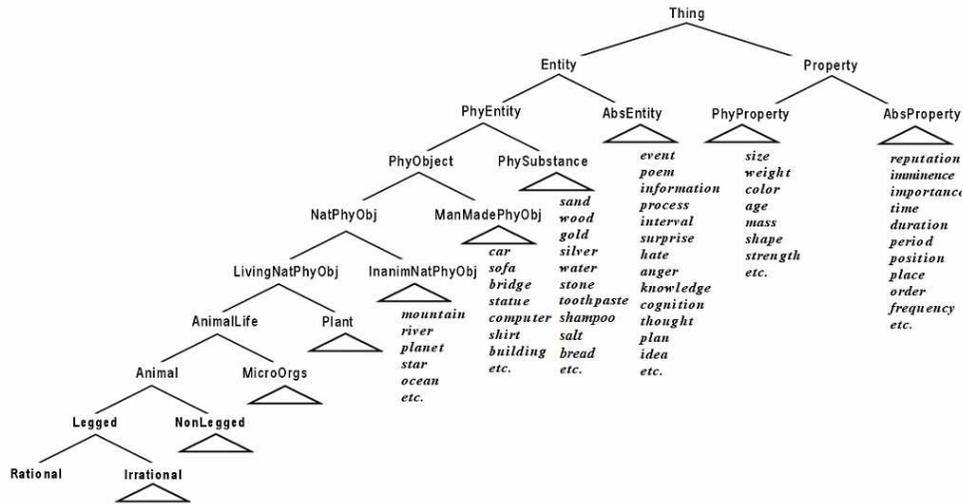

**Figure 4.** A Human is a Physical, LivingThing that is *Formed*, it *Grows*, it *Develops*, it *Moves*, it *Sleeps*, it *Rests*, it (makes sense to say it) *Walks*, *Runs*, *Hears*, *Sees*, *Talks*, *Thinks*, *Reasons*, etc.

## 5. Polysemy and Metaphor

In our approach the occurrence of a verb or an adjective in the hierarchy always refers to a unique sense of that verb or adjective. This has meant that a highly ambiguous verb tends to apply to concepts higher-up in the hierarchy. Moreover, various senses (shades of a meaning) of a verb *v* end-up applying at various levels below *v*. This is illustrated in the small fragment hierarchy shown in figure 5, where we have assumed that we *Make*, *Form*, and *Develop* both an Idea and a Feeling, although an Idea is *Formulated* while a Feeling is *Fostered*. Thus developing, formulating, and forming are considered specific ways of making (that is, one sense of *Make* is *Develop*, or one way of making is developing). While the occurrence of similar senses of verbs at various levels in the hierarchy represents polysemy, the occurrence of the same verb (the same lexeme) at structurally isomorphic places in the hierarchy indicates metaphorical derivations of the same verb. Consider the following:

(8)  *App*(*Run*, LeggedThing)

(9)  *App*(*Run*, Machine)

(10) *App*(*Run*, Show)

(8) through (10) state that we can speak of a legged thing, a machine, and a show running. Clearly, however, these examples involve three different senses of the verb run. It could be argued that the senses of 'run' that are implied by (8) and (10) correspond to a metaphorical derivation of the actual running of natural kinds, the sense implied by (8), as suggested by figure 6.

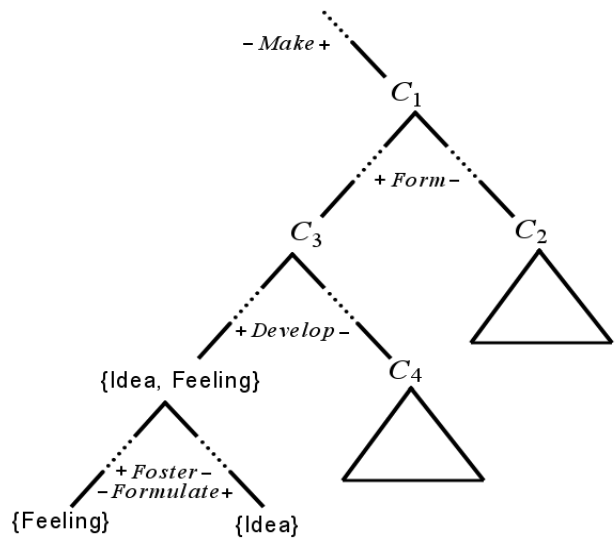

**Figure 5.** An explanation of polysemy.

It is also interesting to note that these metaphorical derivations occur at various levels: first from natural kinds to artifacts; and then from physical to abstract. This is not inconsistent with research on metaphor, such as Lakoff's (1987) thesis that most of linguistic derivations are metaphorical in nature, and that these metaphors are derived from physical concepts (that can all be reduced to a handful of experiential cognitive schemas!)

Note also that the mass/count distinction on the physical side seems to have a mirror image of a mass/count on the abstract side. For example, note the following similarity between water (physical substance) and information (abstract substance, so to speak):

- Water/Information (over)flows
- we filter, process, distill, etc. Water/Information
- Water/Information can be clear and polluted
- we can drown in and be flooded by Water/Information
- a little bit of Water/Information is (still) Water/Information

One interesting aspect of these findings is to further investigate the exact nature of this metaphorical mapping and whether the map is consistent throughout; that is, whether same-level hierarchies are structurally isomorphic, as the case appears to be so far (see figure 6).

## 6. Types vs. Predicates

Although it is far from being complete, in the remainder of the paper we will assume the existence of an ontological structure that reflects our commonsense view of the world and the way we talk about it. As will become apparent, it will not be controversial to assume the existence of such a structure, for two reasons: (*i*) we will only make reference to straightforward cases; and (*ii*) in assuming the existence of such a structure in the analysis of the semantics of the so-called intensional verbs and the semantics of nominal compounds it will become apparent that the analysis itself will in turn shed some light on the nature of this ontological structure. Before we proceed, however, we introduce some additional notation.

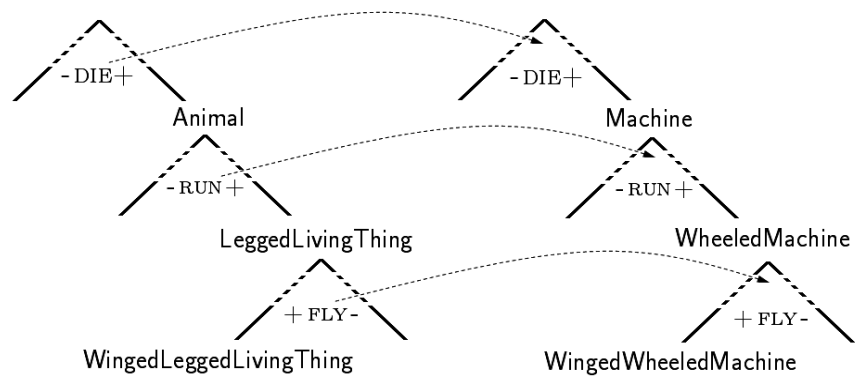

**Figure 6.** Isomorphic structures explaining metaphors.

We shall use a first-order representation where all variables are type annotated. Thus, $x::\text{LivingThing}$ means $x$ is an object of type LivingThing, and $Large(x::\text{Physical})$ means the predicate or property *Large* is true of some $x$ which must be an object of type Physical. We shall write $\text{Human} \prec \text{LivingThing} \prec ... \prec \text{Physical} \prec \text{Thing}$ to state that an object of type Human 'isa' object of type LivingThing, which is ultimately a Physical object, etc. We write $(\exists x :: \text{T})(F(x))$ to state that the property $F$ (which can be a complex logical expression) is true of some (actual) object $x$ of type T; when the property $F$ is true of some 'unique' object $x$ of type T we shall write $(\exists^1 x :: \text{T})(F(x))$ and, finally, we shall write $(\exists^a x :: \text{T})(F(x))$ to state that the property $F$ is true of some object $x$ that only abstractly exists – i.e., an object that need not actually (or physically) exist. Since all variables must be type annotated, a variable in a single scope might receive more than one type annotation, as in the following:

$Artist(x :: \text{Human}) \wedge Old(x :: \text{Physical})$

While *Artist* is a property that can be applied to (or makes sense of) objects that are of type Human, *Old* is a property that makes sense of objects of type Physical. In such an instance some sort of type unification must occur. To illustrate the notion of type unification let us consider the steps involved in the derivation of the meaning of a simple phrase such as 'an old piano':

(11) ⟦*an old piano*⟧
$\quad = (\exists x)(Piano(x) \wedge Old(x))$ \hfill (a)
$\quad = (\exists x)(Piano(x :: \text{Piano}) \wedge Old(x :: \text{Physical}))$ \hfill (b)
$\quad = (\exists x :: Unify(\text{Piano}, \text{Physical}))(Piano(x) \wedge Old(x))$ \hfill (c)
$\quad = (\exists x :: \text{Piano})(Piano(x) \wedge Old(x))$ \hfill (d)
$\quad = (\exists x :: \text{Piano})(Old(x))$ \hfill (e)

In (11a) we have a straightforward translation into first-order logic[4]. However, in our (strongly-typed) approach we require that every variable be annotated (at least once in every scope) with the appropriate type. By the 'appropriate' type we mean the type of object that the predicate (property or relation) applies to (or makes sense of). This is done in (11b), where it was assumed that the predicate *Old* makes sense of (or applies to) objects that are of type Physical. What we now have, however, is an object $x$, which, in a single scope, is considered to be a Piano as well as a Physical object. This necessitates some sort of type unification, as shown in (11c).

Assuming that $\text{Piano} \prec \text{Physical}$ (i.e., assuming our ontology reflects the fact that a Piano is a Physical object), the unification should clearly result in Piano, as given by (11d). Finally, the predicate *Piano* can now be removed without any loss, since it is redundant to state that there is an object of type Piano, of which the predicate *Paino* is true (11e). In the final analysis, therefore, 'an old piano' refers to some object of type Piano, which happens to be *Old*. Note that with this approach there is an explicit differentiation between types and predicates, in that predicates will always refer to properties or relations – what Cochiarella (2001) calls second intension, or logical concepts, as opposed to types (which Cochiarella calls first intension, or ontological concepts)[5]. To appreciate the utility of this representation, consider the steps involved in the derivation of the meaning of 'john is a young professor', where $(\exists^1 j :: \mathsf{Human})$ refers to some unique object $j$ which is of type $\mathsf{Human}$:

(12) ⟦ *john is a young professor* ⟧
$= (\exists^1 j :: \mathsf{Human})(Professor(j :: \mathsf{Human}) \wedge Young(j :: \mathsf{Physical}))$
$= (\exists^1 j :: Unify(\mathsf{Human}, \mathsf{Physical}))(Professor(j) \wedge Young(j))$
$= (\exists^1 j :: \mathsf{Human})(Professor(j) \wedge Young(j))$

---

[4] ⟦*an old piano*⟧ $= \lambda P[(\exists x)(Paino(x) \wedge Old(x) \wedge P(x))]$ is actually the more accurate translation. For simplicity in notation, however, we shall avoid the obvious.

Therefore, while it does not explicitly mention a $\mathsf{Human}$, 'john is a young professor' makes a statement about some unique object $j$ of type $\mathsf{Human}$, which happens to be *Young* and *Professor*. Note, further, that *Professor* in (12) is not a first-intension (ontological) concept, but a second-intension (logical) concept, which is a property of some first-intension concept, namely a $\mathsf{Human}$[6]. Stated yet in other words, what (ontologically) exist are objects of type $\mathsf{Human}$, and not professors, and *Professor* is a mere property that may or may not be true of objects of this type. Moreover, and in contrast with ontological (or first intension) concepts such as $\mathsf{Human}$, concepts such as *Young* and *Professor* are logical concepts in that they are true of a certain object by virtue of some logical expression, as suggested by the following:

(13) $(\forall x :: \mathsf{Physical})(Young(x) \equiv_{df} F_1(x))$
(14) $(\forall x :: \mathsf{Human})(Professor(x) \equiv_{df} F_2(x))$

That is, some $\mathsf{Physical}$ object $x$ is *Young iff* some logical expression $F_1$ is satisfied, and similarly for *Professor*. Furthermore, we suggest that unlike first-intension ontological concepts which tend to be universal and static, second-intension logical concepts tend to be more dynamic and contextual. For example, in

$(\forall x :: \mathsf{Human})(Leader(x) \equiv_{df} F_3(x))$

it can be argued that while $\mathsf{Human} \prec \mathsf{LivingThing}$ is a universal (i.e., shared) fact that can stand the test of time, it is conceivable that the exact nature of the predicate $F_3$ might be susceptible to temporal, cultural, and other contextual factors, depending on what, at a certain point in time, a certain community considers a *Leader* to be[7].

---

[5] Incidentally, Cochiarella (2001) suggests a similar representation where explicit differentiation between types and predicates (relations) is made. Although our starting point was perhaps different, we believe that, ultimately, similar reasons have led to this decision.

[6] Such properties are usually referred to as 'roles'.

## 7. Compositional Semantics Grounded in an Ontology of Commonsense Knowledge

With the machinery developed thus far we are now ready to tackle some challenging problems in the semantics of natural language. In this section we consider the semantics of the so-called intensional verbs, the semantics of nominal compounds and lexical ambiguity.

### 7.1 So-Called Intensional Verbs

Consider the following examples, which Montague (1969) discussed in addressing a puzzle pointed out to him by Quine:

(14) (*a*) ⟦*john painted an elephant*⟧ = $(\exists x)(Elephant(x) \wedge Painted(j, x))$
     (*b*) ⟦*john found an elephant*⟧ = $(\exists x)(Elephant(x) \wedge Found(j, x))$

The puzzle Quine was referring to was the following: both translations admit the inference $(\exists x)(Elephant(x))$ – that is, both sentences imply the existence of some elephant, although it is quite clear that such an inference should not be admitted in the case of (14*a*). In addressing this problem, Montague however discussed the sentence 'John seeks a unicorn'. Using the tools of a higher-order intensional logic, Montague suggested a solution that in effect treats 'seek' as an intensional verb that has more or less the meaning of 'tries to find'. However, this is, at best, a partial solution, since the source of this puzzle is not necessarily in the verb 'seek' nor in the reference to unicorns. Logically speaking, painting, imagining, or even just dreaming about a unicorn does not entail the actual existence of a unicorn – nor does the painting or dreaming about an elephant, or the reader, for that matter. Instead of speaking of intensional verbs, what we are suggesting here is that the obvious difference between (14*a*) and (14*b*) must be reflected in an ontological difference between *Find* and *Paint* in that the extensional type ($e{\rightarrow}(e{\rightarrow}t)$) both transitive verbs are typically assigned is too simplistic. In other words, a much more sophisticated ontology (i.e., a more complex type system) is needed, one that would in fact yield different types for *Find* and *Paint*. One reasonable suggestion for the types of *Find* and *Paint*, for example, could be as follows:

(15) *find* :: $(e_{Animal} \rightarrow (e_{Entity} \rightarrow t))$
(16) *paint* :: $(e_{Human} \rightarrow (e_{Representation} \rightarrow t))$

---

[7] Thanks are due here to an anonymous reviewer who suggested discussing this issue as it pertains to our specific proposal.

That is, instead of the flat type structure implied by $(e \rightarrow (e \rightarrow t))$, what we suggest therefore is that the types of *Find* and *Paint* should reflect our commonsense belief that we can always speak of some Animal that found something (i.e., any Entity whatsoever), and of a Human that painted some Representation. With this background, consider now the translation of 'john found an elephant' which would proceed as follows:

(17) ⟦*john found an elephant*⟧
$= (\exists^1 j :: \text{Human})(\exists x)(\textit{Elephant}(x :: \text{Elephant}) \land \textit{Found}(j :: \text{Animal}, x :: \text{Entity}))$

What we have in (17) is a straightforward translation into first-order logic, where variables are annotated by the appropriate type, and where, as above, by the 'appropriate type' we mean the type of objects that a property or a relation applies to (or makes sense of). Note now that the variables *j* and *x* are annotated, within a single scope, with different types, and thus some type unification must occur, as follows:

(18) ⟦*john found an elephant*⟧
$= (\exists^1 j :: \textit{Unify}(\text{Human}, \text{Animal}))(\exists x)$
$\quad (\textit{Elephant}(x :: \textit{Unify}(\text{Elephant}, \text{Entity})) \land \textit{Found}(j, x))$

Assuming that our ontology reflects the facts that $\text{Human} \prec ... \prec \text{Animal}$ and that $\text{Elephant} \prec ... \prec \text{Entity}$, the type unifications in (18) should result in the following

(19) ⟦*john found an elephant*⟧
$= (\exists^1 j :: \text{Human})(\exists x)(\textit{Elephant}(x :: \text{Elephant}) \land \textit{Found}(j, x))$

Finally, as discussed previously, the predicate *Elephant* can now be removed since its redundant to speak of an object of type Elephant of which the predicate *Elephant* is true, resulting in the following:

(20) ⟦*john found an elephant*⟧ $= (\exists^1 j :: \text{Human})(\exists x :: \text{Elephant})(\textit{Found}(j, x))$

The interpretation of 'John found an elephant' is therefore the following: there is some unique object *j* which is of type Human, and some object *x* which is an Elephant, and such that *j* found *x*. Note, further, that (20) admits the existence of an elephant – that is, if 'John found an elephant' then indeed an actual elephant does exist. However, consider now the interpretation of 'John painted an elephant', which should proceed as follows:

(21) ⟦*john painted an elephant*⟧
$= (\exists^1 j :: \text{Human})(\exists x)(\textit{Elephant}(x :: \text{Elephant}) \land$
$\quad \textit{Painted}(j :: \text{Animal}, x :: \text{Representation}))$

As in (17), type unification for the variables *j* and *x* must now occur. Also, as in (18), *Unify*(Human, Animal) should also result in Human. Unlike the situation in (18), however, resolving the type the variable *x* must be annotated with is not as simple. Since the types Elephant and Representation are not related by the 'isa' relationship, we are in fact referring to two genuinely different types and some relation between them, say *RepresentationOf*[8], with the caveat that one of these objects need not actually exist. What we suggest therefore is the following:

(22) ⟦ *john painted an elephant* ⟧
= $(\exists^1 j :: \mathsf{Human})(\exists^a x :: \mathsf{Elephant})(\exists y :: \mathsf{Representation})$
$(RepresentationOf(y,x) \land Painted(j,y))$

Essentially, therefore, 'john painted an elephant' roughly means 'join made a representation of some object (that need not physically exist), an object that we call an elephant'. Note now that if 'john painted an elephant' than what exists is a Representation of some object of type Elephant[9]. Thus, while (22) admits the existence of some Representation, (22) only admits the abstract existence of some object we call an Elephant.

Finally it should point out that while the interpretation of 'John painted an elephant' given in (22) allows one to make the right inference, namely that a representation and not an elephant is what actually exists, one should also be able to make several other inferences. This would actually require a more elaborate event-based representation. For example, consider the following:

⟦John bought an old piano⟧
= $(\exists e :: \mathsf{BuyingEvent})(\exists^1 j :: \mathsf{Human})(\exists x :: \mathsf{Piano})$
$(Old(x) \land Agent(e,j) \land Object(e,x))$

That is, 'John bought an old piano' essentially says there is some unique object *j* which is of type Human, some object *x* of type Piano, such that *x* is *Old*, and such that *j* and *x* are involved in some BuyingEvent *e* as follows: *j* is the agent of the event and *x* is the object. Assuming that Piano 'isa' MusicalInst and BuyingEvent 'isa' PurchasingEvent; then the following inferences, among others, can be made:

⟦John bought an old piano⟧
= $(\exists e :: \mathsf{BuyingEvent})(\exists^1 j :: \mathsf{Human})(\exists x :: \mathsf{Piano})$
$(Old(x) \land Agent(e,j) \land Object(e,x))$

---

[8] In the ontology, *RepresentationOf*(*x,y*) would actually be defined between a Representation and a type higher-up in the hierarchy, e.g. Entity.

[9] The point of this example will perhaps be made more acutely if 'elephant' was replaced by 'unicorn'.

$$\Rightarrow (\exists e :: \mathsf{PurchasingEvent})(\exists^1 j :: \mathsf{Human})(\exists x :: \mathsf{Piano})$$
$$(Old(x) \land Agent(e, j) \land Object(e, x))$$
$$\Rightarrow (\exists e :: \mathsf{BuyingEvent})(\exists^1 j :: \mathsf{Human})(\exists x :: \mathsf{MusicalInstr})$$
$$(Old(x) \land Agent(e, j) \land Object(e, x))$$
$$\Rightarrow etc.$$

That is, if 'John bought an old piano' then one must be able to infer that 'John purchased a piano', 'John bought an old musical instrument', etc.

### 7.2 The Semantics of Nominal Compounds

The semantics of nominal compounds have received considerable attention by a number of authors, most notably (Kamp & Partee, 1995; Fodor & Lepore, 1996; Pustejovsky, 2001), and to our knowledge, the question of what is an appropriate semantics for nominal compounds has not yet been settled. Recall that the simplest (extensional) semantic model for simple nominal constructions is that of conjunction (or intersection) of predicates (or sets). For example, assuming that $Red(x)$ and $Apple(x)$ represent the meanings of red and apple, respectively, then the meaning of a nominal such as red apple is usually given as

(23)  $[\![red\ apple]\!] = \{x \mid Red(x) \land Apple(x)\}$

What (23) says is that something is a red apple if it is red and apple. This simplistic model, while seems adequate in this case (and indeed in many other instances of similar ontological nature), clearly fails in the following cases, both of which involve an adjective and a noun:

(24) *former senator*

(25) *fake gun*

Clearly, the simple conjunctive model, while seems to be adequate for situations similar to those in (23), fails here, as it cannot be accepted that something is a former senator if it is former and senator, and similarly for (25). Thus, while conjunction is one possible function that can be used to attain a compositional meaning, there are in general more complex functions that might be needed for other types of ontological categories. In particular, what we seem to have here is something like the following:

(26) *a)*  $[\![red\ apple]\!] = \{x \mid Red(x) \land Apple(x)\}$
    *b)*  $[\![former\ senator]\!] = \{x \mid WasButIsNotNowASenator(x)\}$
    *c)*  $[\![fake\ gun]\!] = \{x \mid LooksLikeButIsNotActuallyAGun(x)\}$

The fact that every adjective-noun combination seem to suggest a different compositional function have led some authors to argue against compositionality (e.g., Lahav, 1989).

However, it would seem in fact that there might be just a handful of templates of compositional functions for a number of ontological categories. Consider for example the following reasonable definitions for *Fake* and *Former*:

(27) $(\forall x :: \mathsf{Human})(Former(x) \equiv_{df} \lambda P[(\exists t)((t < now) \wedge P(x,t) \wedge \neg P(x, now))])$

(28) $(\forall x :: \mathsf{Artifact})(Fake(x) \equiv_{df} (\exists y : \mathsf{PhyEntity})(\neg IsA(x, y) \wedge Similar_A(x, y)))$

What (27) says is the following: a certain $x$ is a *Former P* iff $x$ was a *P* at some point in time in the past and is not now a *P*, where *P* is some property which applies to objects of type Human. On the other hand, what (28) says is that a certain object $x$, which must be of type Artifact, is a *Fake y*, which must be an object of type PhyEntity, *iff* $x$ is not actually a $y$ but is similar, with regard to some property $A$, to $y$.

First, it is interesting to note here that the intension of *Fake* and *Former* was in one case represented by recourse to possible worlds semantics (27), while in (28) the intension uses something like structured semantics, assuming that $Similar_A(\mathsf{x,y})$ which is true of some $x$ and some $y$ if $x$ and $y$ share a number of important features, is defined. What is interesting in this is that it suggests that possible-worlds semantics and structured semantics are not two distinct alternatives to representing intensionality, as has been suggested in the literature, but that in fact they should co-exist. Note further that the proposed meaning of *Fake* given in (28) suggests that fake expects a concept which is of type Physical, and thus something like *fake idea*, or *fake song*, etc., should sound meaningless from the standpoint of commonsense[10].

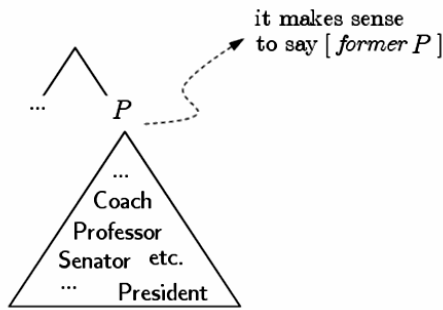

**Figure 7.** Roles that share the same behavior

Second, the proposed meaning of 'former' given in (27) suggests that former expects a property which has a time dimension, i.e. is a temporal property. Finally, we should note that the goal of this type of analysis is to discover the ontological categories that share the same behavior. For example, conjunction, which as discussed above is one possible function that can be used to attain a compositional meaning, seems to be adequate for all nominal constructions of the form [A N] where A is a PhyProperty (such as Color, Weight, Size, etc.) and N is a PhyObject (such as Car, Person, Desk, etc.), as expressed in (29).

---

[10] One can of course say *fake smile*, but this is clearly another sense of *fake*. While *fake gun* refers to a gun (an Artifact) that is not real, *fake smile* refers to a dishonest smile, or a smile that is not genuine.

(29)  $[\![A\ N]\!] = \{x \mid A_{PhyProperty}(x) \wedge N_{PhyObject}(x)\}$

Similarly, an analysis of the meaning of 'former', given in (27), suggests that there are a number of ontological categories that seem to have the same behavior, and could thus replace P in (27), as implied by the fragment hierarchy shown in figure 7. Finally it should be noted here that (29) simply states that some adjectives are intersective, although it does not say anything about the meaning of any particular adjective. While this is not our immediate concern, such concepts are assumed to be defined by virtue of a logical expression. To do so, we assume first the existence of a predicate $Typical_A^P(x :: \mathsf{T})$, which is used to state that an object of type T is a typical P as far as some attribute A is concerned, where the typicality of a certain object regarding some attribute is assumed to be defined by virtue of some logical expression $\alpha$, as $(\forall x :: \mathsf{T})(Typical_A^P(x) \equiv_{df} \alpha)$. For example, the following is an expression that defines the typical age of a gymnast, $Typical_{Age}^{Gymnast}(x)$:

$$(\forall x :: \mathsf{Human})(Typical_{Age}^{Gymnast}(x) \equiv_{df} (Age(x,a) \wedge (n \leq a \leq m)))$$

Consider now the definition of an adjective such as 'old', as it relates to the age of objects of type Human:

$$(\forall x :: \mathsf{Human})(Old_P(x) \equiv_{df} \lambda P[P(x) \wedge (\exists y :: \mathsf{Human})(P(y) \wedge Typical_{Age}^P(y)$$
$$\wedge Age(x,ax) \wedge Age(y,ay) \wedge (ax \gg ay))])$$

What the above is saying is the following: some object x of type Human is an *Old P*, *iff* its *Age* is larger than the *Age* of another object, y, which has a typical *Age* as far as P objects are concerned. It would seem, then, that the meaning of such adjectives is tightly related to some attribute (large/size, heavy/weight, etc.) of the corresponding concept. Furthermore, it would seem that some adjectives are context-dependent in two respects: the types of objects they apply to (or makes sense of) as well as the property or relation that the adjective is to modify. That is, 'old' in 'old person' is quite different from 'old' in 'old piano'. Furthermore, a certain object of type Human, say, can be an 'old P' and a 'young Q' at the same time. For example, consider the following:

$[\![john\ is\ an\ old\ gymnast]\!] = (\exists^1 j :: \mathsf{Human})(Gymnast(x) \wedge Old_{Gymnast}(x))$

Given that 'John is an old gymnast', and, for example, 'John is a professor', one would not, in our representation, conclude that 'John is an old professor', since John is an old gymnast while it might very well be the case that as far professors go, John is quite young.

## 7.3 Compositional Semantics of Nominal Compounds

While the semantics of [*Adj Noun*] constructions can be problematic, it is the semantics of nominal compounds in the case of noun-noun combinations that have traditionally posed a challenge to the whole paradigm of compositional semantics. The difficulty in analyzing the meaning of noun-noun combinations is largely due to the multitude of possible relations that are usually implicit between the two nouns. For example, consider the following:

(30) 〚*brick house*〛 = {$x \mid x$ is a House that is *MadeOf* Brick}
(31) 〚*dog house*〛 = {$x \mid x$ is a House that is *MadeFor* a Dog}

That is, while a *brick house* is a house 'made of' brick, a *dog house* is a house that is 'made for' a dog. It would seem, then, that the relation implicitly implied between the two nouns differs with different noun-noun combinations. However, assuming the existence of a strongly-typed ontology might result in identifying a handful of patterns that can account for all noun-noun combinations. As shown in the fragment hierarchy of figure 8, it would seem that *MadeOf* is the relation implicit between all [$N_1$ $N_2$] combinations whenever $N_1$ is a Substance and $N_2$ is an Artifact, which expressed more formally in (32).

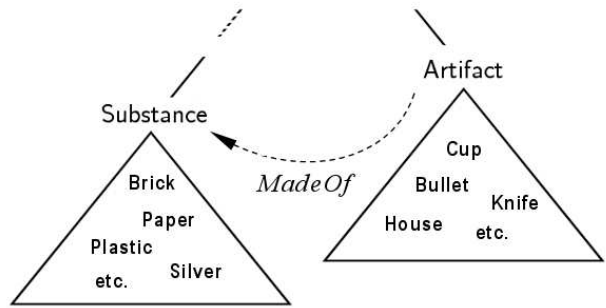

**Figure 8.** Relations between ontological categories

(32)  〚$N_{\text{Substance}} N_{\text{Artifact}}$〛 = $\{x :: \text{Artifact} \mid (\exists y :: \text{Substance})(MadeOf(x, y))\}$

The following is an example instance of (32), denoting the meaning of *brick house*, where it is assumed that our ontology reflects the fact that House $\prec ... \prec$ Artifact  and that Brick $\prec ... \prec$ Substance :

(33) 〚*brick house*〛 = $\{x :: \text{House} \mid (\exists y :: \text{Brick})(MadeOf(x, y))\}$

Note, further, that specific instances of Substance and specific instances of Artifact might require the specialization of the relation suggested in (32). For example, while Knife which is an Artifact, combines with a raw Substance, such as Plastic, Bronze, Wood, Paper, etc. with the relation *MadeOf*, Knife as an Instrument combines with a FoodSubstance, such as Bread, with the relation *UsedFor*, and similarly for Coffee and Mug, and Cereal and Box, as follows:

(34) $[\![ N_{\text{FoodSubstance}} N_{\text{Instrument}} ]\!] = \{x :: \text{Instrument} \mid (\exists y :: \text{FoodSubstance})(UsedFor(x, y))\}$

Although we will not dwell on such details here, we should point out here that since the purpose of an object of type Instrument (and more specifically, a Tool) is to be used for something, the specific type of usage would in turn be inferred from the specific Instrument/Tool (e.g., *cutting* in the case of a Knife, *holding* in the case of Mug, etc.)

**7.4 Lexical Disambiguation as Type Inferencing**

First let us suggest the following types for the transitive verbs *marry* and *discover*:

(35) $marry :: (e_{\text{Human}} \to (e_{\text{Human}} \to t))$
(36) $discover :: (e_{\text{Animal}} \to (e_{\text{Entity}} \to t))$

That is, we are assuming that one sense of the verb *marry* is used to refer to an object of type Human that may stand in the 'marry' relationship to another object which is also of type Human; and that it makes sense to speak of an object of type Animal that discovered an object of type Entity. Consider now the following, where $Star(x :: \{\text{Human}, \text{Star}\})$ is used to refer to the fact that *Star* is a predicate that applies to, among possibly some others, an object of type Human or an object of type Star (which is a subtype of CelestialBody):

(37) $[\![\, john\ married\ a\ star\,]\!]$
$= (\exists^1 j :: \text{Human})(\exists x)(Star(x :: \{\text{Human}, \text{Star}\}) \land Married(j :: \text{Human}, x :: \text{Human}))$

As usual, since $x$ is annotated with more than one type in a single scope, some type unification must occur. The unification between Star (the CelestialBody) and Human will fail, however, leaving one possible meaning for (37):

(38) $[\![\, john\ married\ a\ star\,]\!]$
$= (\exists^1 j :: \text{Human})(\exists x :: \text{Human})(Star(x) \land Married(j, x))$

Note, therefore, that 'star' in 'John married a star' was translated to a property of an object of type Human, rather than an ontological type, such as Star, which is a subtype of CelestialBody. However, consider now the following:

(39) $[\![\, john\ discovered\ a\ star\,]\!]$
$= (\exists^1 j :: \text{Human})(\exists x)(Star(x :: \{\text{Human}, \text{Star}\})$
$\quad \land Discovered(j :: \text{Animal}, x :: \text{Entity}))$

In this case both type unifications are possible, as Entity unifies with both a Human and a Star (and of course Animal trivially unifies with Human), resulting in two possible meanings, in which 'star' is translated into a property of an object of type Human, and into an ontological object referring to a celestial body, as given in the following:

(40) ⟦*john discovered a star*⟧
$\Rightarrow (\exists^1 j :: \mathsf{Human})(\exists x :: \mathsf{Human})(Star(x) \land Discovered(j,x))$
$\Rightarrow (\exists^1 j :: \mathsf{Human})(\exists x : \mathsf{Star})(Discovered(j,x))$

Lexical disambiguation will clearly not always be as simple, even with a rich ontological structure underlying the various lexical items. For one thing, several lexical items might be ambiguous at once, as the following example illustrates:

(41) ⟦*john is playing bridge*⟧
$= (\exists^1 j :: \mathsf{Human})(\exists x)\ (Bridge(x :: \{\mathsf{CardGame}, \mathsf{Structure}\})$
$\qquad \land Playing(j :: \mathsf{Animal}, x :: \{\mathsf{Game}, \mathsf{Instrument}\}))$

Here it was assumed that 'bridge' can refer to a Structure or to a CardGame, while 'playing' can be a relation that holds between an object of type Animal and a Game, or an object of type Human and an Instrument. While there are potentially four possible readings for (41) that are due only to lexical ambiguity, CardGame and Game is the only successful type unification, resulting in the following:

(42) ⟦*john is playing bridge*⟧
$= (\exists^1 j :: \mathsf{Human})(\exists x :: \mathsf{CardGame})\ (Bridge(x) \land Playing(j,x))$

Finally it must be noted that in many instances the type unification, while it might result in more than one *possible* unification, one of which, might be more *plausible* than the others. That, however, belongs to the realm of pragmatics, and requires type information form larger linguistic units, perhaps at the level of a paragraph. While we cannot dwell on this issue here, the point of the above discussion, as in the case in our discussion of nominal compounds, was to simply illustrate the utility of a strongly-typed ontological structure that reflects our commonsense view of the world in tackling a number of challenging problems in the semantics of natural language.

## 8. Towards a Meaning Algebra

If Galileo was correct and mathematics is the language of nature, then Richard Montague (see the paper on ELF in (Thomasson, 1974)), is trivially right in his proclamation that there is no theoretical difference between formal and natural languages. Moreover, if Montague is correct, then there should exists a formal system, much like arithmetic, or any other algebra, for concepts, as advocated by a number of authors, such as Cresswell (1973) and Barwise (1989), among others. What we are arguing for here is a formal system that explains how concepts of various types combine, forming more complex concepts. To illustrate, consider the following:

(43) *a*) *artificial* :: NatPhyObj $\rightarrow$ ManMadePhyObj
    *b*) *flower* :: Plant $\prec$ ... $\prec$ LivingThing $\prec$ ... $\prec$ NatPhyObj $\prec$ ... $\prec$ Thing
    *c*) *artificial flower* :: ManMadePhyObj

What the above says is the following: *artificial* is a property that applies on an object of type NatPhyObj returning in an object of type ManMadePhyObj (43*a*); a *flower* is a Plant, which is a LivingThing, which in turn is a NatPhyObj (43*b*); and, finally, an artificial flower is a ManMadePhyObj (43*c*). Therefore, 'artificial *c*', for some NatPhyObj *c*, should in the final analysis have the same properties that any other ManMadePhyObj has. Thus, while a *flower*, which is of type Plant, and is therefore a LivingThing, grows, lives and dies like any other LivingThing, an 'artificial flower', and like any other ManMadePhyObj, is something that is manufactured, does not grow, does not die, but can be assembled, destroyed, etc.

The concept algebra we have in mind should also systematically explain the interplay between what is considered commonsense at the linguistic level, type checking at the ontological level, and deduction at the logical level (recall figure 1). For example, the concept 'artificial car', which is a meaningless concept from the standpoint of commonsense, is ill-typed since Car is an ManMadePhyObj, which does not unify with NatPhyObj. The concept 'former father', on the other hand, which is also a meaningless concept from the standpoint of commonsense, escapes type-checking since *Father* is a property that applies to objects of type Human, as expected by the meaning of 'former'. However, the reader can easily verify that the meaning of 'former' suggested in (27) and the following,

$$(\forall x :: \text{Human})((\exists t_1)(Father(x, t_1) \supset (\forall t_2)((t_2 > t_1) \supset Father(x, t_2))))$$

which states a temporal property about the concept of 'father', namely that once an object of type Human is father then they are always a father, lead to a logical contradiction. What we envision, therefore, is a logic that has content, and ontological content in particular, and where linguistic expressions that do not confirm with our commonsense view of the world, are either caught at the type-checking level, or, if it escapes type- checking, is caught at the logical level.

## 9. Concluding Remarks

In this paper we argued for and presented a new approach to the systematic design of ontologies of commonsense knowledge. The method is based on the basic assumption that "language use" can guide the classification process. This idea is in turn rooted in Frege's principle of Compositionality and is similar to the idea of type inference in strongly-typed, polymorphic programming languages.

The experiment we conducted shows this approach to be quite promising as it seems to simultaneously provide for an adequate solution to a number of problems in the semantics of natural language. Admittedly, however, much of what we presented here is work in progress, more so than a final result, and much work remains to be done. In particular, we are in the process of automating the process described in section 4 by using a corpus analysis that would generate sets of concepts for which adjectives and verbs may or may not apply. Another interesting aspect of this work is identifying the top-level categories that share the same behavior, leading to the identification of a number of template compositional functions, as those given in (26) and (27), a step that is essential in our quest for a meaning algebra that is grounded in a strongly-typed ontology that reflects our commonsense view of the world and the way we talk about it in ordinary language.